\documentclass[letterpaper, 10 pt, conference]{ieeeconf}  

\IEEEoverridecommandlockouts                              

\usepackage[hidelinks]{hyperref}
\usepackage{todonotes}

\newcommand{\etal}{\emph{et al.}}
\newcommand{\algname}{VLM-Vac}

\title{\LARGE \bf
\algname: Enhancing Smart Vacuums through VLM Knowledge Distillation and Language-Guided Experience Replay
}

\author{Reihaneh Mirjalili$^{1}$, Michael Krawez$^{1}$, Florian Walter$^{1}$ and Wolfram Burgard$^{1}$
\thanks{$^{1}$All authors are with the Department of Engineering, University of
Technology Nuremberg, Germany.
        }%
}
\usepackage{graphicx}
\usepackage[export]{adjustbox}
\usepackage{amsmath}

\usepackage[T1]{fontenc}
\usepackage{array}
\usepackage{booktabs}

\usepackage{pbox}
\begin{document}
\maketitle
\thispagestyle{empty}
\pagestyle{empty}
\begin{abstract}
In this paper, we propose \algname, a novel framework designed to enhance the autonomy of smart robot vacuum cleaners. Our approach integrates the zero-shot object detection capabilities of a Vision-Language Model (VLM) with a Knowledge Distillation (KD) strategy. By leveraging the VLM, the robot can categorize objects into actionable classes---either to avoid or to suck---across diverse backgrounds. However, frequently querying the VLM is computationally expensive and impractical for real-world deployment. To address this issue, we implement a KD process that gradually transfers the essential knowledge of the VLM to a smaller, more efficient model. Our real-world experiments demonstrate that this smaller model progressively learns from the VLM and requires significantly fewer queries over time. Additionally, we tackle the challenge of continual learning in dynamic home environments by exploiting a novel experience replay method based on language-guided sampling.
Our results show that this approach is not only energy-efficient but also surpasses conventional vision-based clustering methods, particularly in detecting small objects across diverse backgrounds.

\end{abstract}

\section{Introdcution}
Robotic systems are increasingly integrated into our everyday life, playing critical roles in various domains including manufacturing, logistics, transportation, entertainment, and home automation. Despite their broad adoption, domestic robots in particular face persistent challenges due to the complexity and diversity of everyday environments and tasks. While advanced algorithms and sensors enhance these robots in terms of navigation and operation, many domestic robots still struggle with perception and decision-making tasks.
Smart vacuum cleaners, for instance, encounter various challenges in this regard. Naively covering the entire floor area can be problematic as it does not guarantee cleaning, might spread liquid or sticky substances, and risks sucking up valuable items.
Recent advances in computer vision and machine learning have aimed to address these challenges. Data-driven methods for dirt detection have been proposed in several works ~\cite{bormann2020dirtnet, canedo2021deep, yun2022deep}. However, these approaches rely heavily on manually annotated datasets, which can be costly and limit their practical applicability in real-world scenarios.

\begin{figure}[t!]
\centering
\includegraphics[clip,trim=0cm 8cm 10cm 0cm,width=0.95\linewidth]{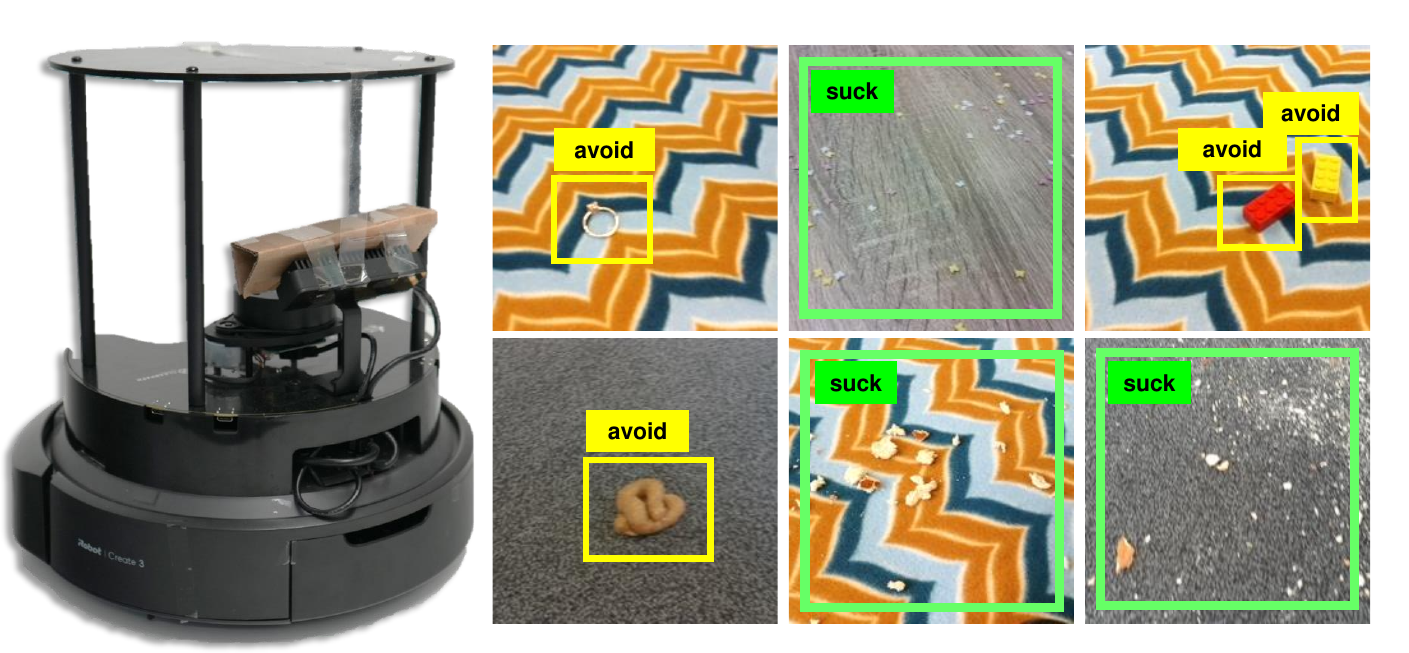}  
\caption{Overview of our smart robotic vacuum cleaner system. The TurtleBot 4 platform, shown on the left, is used in our experiments and resembles a robotic vacuum cleaner. The images on the right, captured by the robot's camera, illustrate our system's real-time detection of ``suck'' or ``avoid'' actions.}
\label{fig:coverpic}
\end{figure}

Large Language Models (LLMs) and VLMs have shown impressive capabilities in image understanding and common sense reasoning, benefiting from extensive training data.
However, the practical deployment of these models is hindered by their high inference costs, whether due to energy consumption for self-hosted models or expenses related to commercial services. 
Addressing these challenges requires innovative methods that balance performance and efficiency.

In this paper, we introduce \algname{}, a novel approach that leverages multimodal foundation models to enable smart vacuum cleaners to autonomously decide whether to clean or avoid an area. Our system integrates knowledge distillation (KD) with language-based continual learning to enhance functionality. Specifically, we fine-tune YOLOv8n~\cite{yolov8_ultralytics} using bounding boxes labeled with actionable categories: ``suck'' or ``avoid''. These bounding boxes and labels are generated through the combination of a VLM and an open-vocabulary object detection model. During operation, when the robot encounters unfamiliar objects or floor patterns, it queries the VLM. This new data instance is then stored and used to fine-tune the action-based object classification model. Over time, this approach reduces the number of queries to the VLM, thus improving system efficiency\footnote[2]{Video available at 
\href{https://tinyurl.com/yakh5vrf}{https://tinyurl.com/yakh5vrf}.
}. 

\textbf{In summary, we make the following contributions: }
\begin{enumerate} 
    \item We propose \algname{}, a novel framework that enhances the functionality of smart vacuum cleaners by leveraging VLMs.
    \item We implement a knowledge distillation process to transfer essential knowledge from a computationally expensive VLM to a smaller, more efficient model.
    \item We leverage a novel language-guided experience replay method for continual learning in dynamic environments.
    \item We present a new dataset of images captured by a TurtleBot~4 robot, representing a vacuum cleaner in a household setting, and thoroughly evaluate our approach with real robot data.
    \item We will make our dataset and code publicly available.

\end{enumerate}

\section{Related Work}
\label{sec:related_work}
Detecting and classifying dirt on floors is an active research topic in cleaning robotics, with earlier approaches relying on saliency detection~\cite{bormann2013autonomous}, spectral analysis~\cite{milinda2017mud}, or regular pattern analysis for floor-waste separation~\cite{ramalingam2018vision}. Later works employ data-driven methods~\cite{bormann2020dirtnet, canedo2021deep, yun2022deep, singh2023vision, guan2022dirt}, with many of them utilizing a version of the YOLO-architecture. These models are trained on dedicated datasets which are small compared to modern VLM training data sizes and thus do not support detection in an open-world scenario. Novelty detection using a CNN has been proposed as a method for floor inspection~\cite{grunwald2018optical}. However, it does not classify detected items or stains.
Recent advancements include Xu \etal~\cite{xu2024sweepmm}, who propose a multi-modal dataset tailored for household sweeping robots and use it to fine-tune an LLM for several cleaning-related downstream tasks. For instance, the model supports open-world object detection and provides cleaning recommendations according to user recommendations. In contrast to our work, the authors are not targeting model size reduction and the model does not adapt to the environment over time.

Continual learning, becomes crucial for robots functioning in real-world environments. As these models need to adapt to evolving data distributions without forgetting prior knowledge. 
Techniques such as Elastic Weight Consolidation (EWC)~\cite{kirkpatrick2017overcoming} address catastrophic forgetting by penalizing changes in model parameters that are important for the previously learned task. Peng \etal~\cite{peng2023diode} observe that the performance of regularization methods for continual learning like EWC deteriorates with more new tasks learned by a model. As one possible cause, the authors identify the progressively decreasing number of free parameters which are not used for previously learned tasks. To elevate this problem, they propose to extend the model with new parameters for each new task. Liu \etal~\cite{liu2020incdet} link poor performance of EWC on continual object detection to exploding gradients caused by the quadratic term in the EWC loss and missing bounding boxes of old classes in new training data. The authors propose to use Huber regularization to stabilize training and to deploy the previous model version to generate bounding boxes of old classes.
Liu~\etal~\cite{liu2020multi} address continual learning in object detection by combining a parameter regularization method with experience replay. In particular, they add a so-called Attentive Feature Distillation term to the loss function, consisting of a bottom-up and a top-down part. The bottom-up term selects and preserves parameters relevant for previous tasks. The top-down attention is computed from the intersection of ground truth bounding boxes and region proposals of the model and aims at distilling foreground objects. Finally, the authors propose a memory buffer sampling scheme which accounts for the number of bounding boxes per training image.
Several studies analyse different continual learning approaches in various scenarios.
Wang \etal~\cite{wang2021wanderlust} create a dataset and benchmark for online continual object detection. They further evaluate several continual learning baselines on the benchmark, including EWC~\cite{kirkpatrick2017overcoming} and iCaRL~\cite{rebuffi2017icarl}. The latter shows the best performance on the baseline, though lagging behind a model trained offline on all object categories. 
Kalb \etal~\cite{kalb2021continual} evaluate several continual learning baselines in the field of semantic segmentation. Their findings suggest that KD approaches that employ a previously trained model as a teacher for the new model perform better in class-incremental tasks, while replay-based methods are better suited for domain-incremental learning. In our work, we rely on experience replay, since we expect possible domain changes like newly discovered floor patterns.
Building on these advancements, integrating language-based descriptors and experience replay methods can significantly enhance continual learning in dynamic environments. For instance, Stephan~\etal\ \cite{stephan2024text} argue that language can serve as a robust image descriptor and propose a text-guided image clustering method. Similarly, El Banani, Desai, and Jonson \cite{el2023learning} propose sampling training image pairs based on their caption similarity which captures semantic rather than merely visual similarities.   

With the recent emergence of foundation models, VLMs opened the door for open-vocabulary object detection and boosted visual perception \cite{Lan-grasp}, \cite{fm-loc}.
However, the size of these models is a prohibiting factor for many applications. Therefore significant efforts~\cite{xu2024survey} were put into KD, i.e., training approaches that use a large expert model to guide the training of a small and efficient one. Gu \etal~\cite{guopen} trains a Mask R-CNN architecture as an open-vocabulary detector by aligning the object proposal embeddings with the embeddings of the teacher model. KD was also used by Ma \etal~\cite{ma2022open} to transfer class-level and instance-level knowledge from a teacher to an efficient network for open-vocabulary object detection.
A number of works applied KD to train a model for a specific downstream task.
Sumers \etal~\cite{sumers2023distilling} leverage a VLM to label trajectories and goal states to teach a robotic agent object categories and their attributes. Yang \etal~\cite{yang2024embodied} use a pre-trained LLM to guide the training of a VLM-based agent. Building upon these existing methods, our approach integrates KD with language-based continual learning to enhance the adaptability and efficiency of smart vacuum cleaner robots in diverse and dynamic environments.

\section{Method Description}
\label{sec:approach}
In this section, we introduce the details of \algname{}. Our approach consists of two main components. We first describe the distillation process from the VLM, GPT-4o in our case, to the smaller model, YOLOv8n. We then proceed to detail our continual learning method, which utilizes language-based experience replay. 

\begin{figure*}[ht]
\centering
\includegraphics[clip,trim=0cm 3cm 0cm 0cm,width=0.9\linewidth]{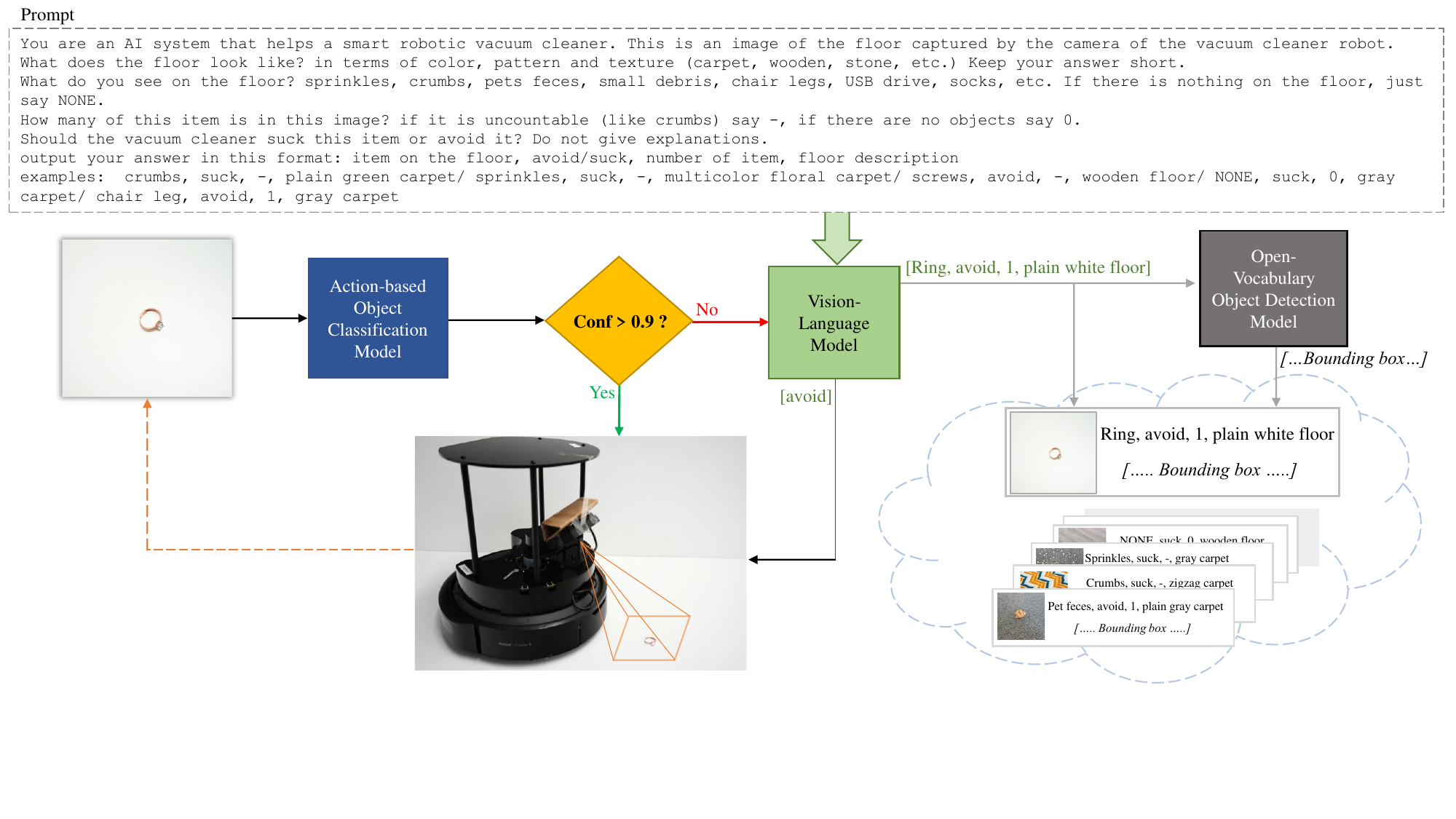}  
\caption{\algname{} in a nutshell: We distill relevant knowledge from a Vision-Language Model (VLM) into a compact action-based object detector. The smaller model queries the VLM whenever it encounters an unfamiliar object or background. The new image, its text description from the VLM, and corresponding bounding boxes from the open vocabulary object detector are stored in the experience pool and later used for training the smaller model. Over time, the smaller model learns from these interactions, adapting to its specific environment and thus reducing the need for VLM queries. }
\label{fig:overview}
\end{figure*}

\subsection{Knowledge Distillation for Action-based Object Classification}
VLMs are ideal tools to extract semantic understanding about items and flooring patterns in a zero-shot manner.
However, since continuously querying a VLM is computationally expensive and thus impractical for real-world applications, we employ a KD technique to transfer the relevant knowledge from the VLM to a smaller, more efficient object detection model, specifically YOLOv8n \cite{yolov8_ultralytics} in our approach.
Based on this technique, when YOLOv8n encounters an unfamiliar object that it cannot classify with a defined confidence threshold, it queries the VLM for guidance and gradually learns from these interactions. 
The general process, along with the prompt \(\mathrm{p}\), is depicted in \autoref{fig:overview}: The image \( I \) captured by the robot's camera is fed to the GPT-4o VLM along with the specified prompt to get the text description \(\mathrm{t}(c, q, a, f)\) which provides information on the item category \( c \), quantity \( q \), action class \( a \) and floor type \( f \):
\begin{equation}
\mathrm{t}(c, q, a, f) = \text{VLM}(\mathrm{p}, I)
\end{equation}
Additionally, we use OWL-ViT \cite{owl-vit} for grounding the item label detected by GPT-4o in the image through bounding boxes. A new image \(I^{new}\) is stored in what we call an \textit{experience pool} \( \mathcal{E} \) along with its text description \(\mathrm{t}^{new}\) and corresponding bounding boxes \( \mathrm{B}^{new} \).

After each day of operation, YOLOv8n is subsequently trained on an \textit{experience buffer} \(\mathcal{B}\):
\begin{equation*}
\mathcal{B}  = \{(I_1, a_1, \mathrm{B}_1),\dots, (I_N, a_N, \mathrm{B}_N) \}
\end{equation*}
\(\mathcal{B}\) is sampled from the experience pool \( \mathcal{E} \) based on a selection strategy that focuses on choosing a balanced mix of image instances along with their corresponding bounding boxes and action classes. This selection process is crucial, and we will discuss it in greater detail in the following subsection.

Using this approach, the frequency of VLM queries is expected to decrease over time due to the effective distillation of knowledge into YOLOv8n. Consequently,  the vacuum cleaner will gradually adapt to its surrounding environment and, as a result, become more efficient and responsive.

\subsection{Continual Learning with Language-Based Experience Replay}
As previously discussed, a critical aspect of our approach is the data selection strategy for fine-tuning YOLOv8n. One straightforward method is \textit{cumulative training}, which involves training on all the accumulated data.
While this method exhibits a strong performance, it is computationally expensive and thus not suitable for practical resource-constrained applications like mobile robots.
An alternative approach is \textit{naive fine-tuning} of the model solely on new data, e.g. all samples acquired throughout the preceding day.
Although this is more cost effective than cumulative learning, it presents significant challenges: Vacuum cleaner robots operate in dynamic domestic environments, where the collected data can vary significantly over time. It is therefore critical to retain previously learned knowledge while adapting to changes in order to improve over time. For instance, the robot may encounter specific items, such as pet waste, infrequently or operate in different rooms with varying floor patterns on different days. Learning only from the most recent experiences can lead to what is known as \emph{catastrophic forgetting} \cite{mccloskey1989}, where previously acquired knowledge is eroded, posing a substantial challenge in continual learning scenarios. 

To tackle this issue, we take inspiration from Stephan~\etal\cite{stephan2024text} and exploit a novel \textit{language-based experience replay} method. As mentioned earlier, whenever the robot encounters an unfamiliar object with low confidence score, it queries the VLM and stores the new image \( I^{new} \), along with its text description \( \mathrm{t}^{new} \), quantity \( q^{new} \), action class \( a^{new} \) and floor pattern \( f^{new} \) in the experience pool \( \mathcal{E} \). We exploit language-based experience replay to group similar items---considering both the background and the item category---into the same cluster. 
To achieve this, we extract a language embedding \( e^{new} \) based on the item category \( c^{new} \), quantity \( q^{new} \), action class \( a^{new} \) and floor patter \( f^{new} \):

\begin{equation}
e^{new} = \text{Embedding}(c^{new}, q^{new}, a^{new}, f^{new})
\end{equation}

We then apply \(k\)-means clustering to group similar experiences together based on these embeddings:
\begin{equation}
\min_{\mu} \sum_{k=1}^{K} \sum_{e \in \mathcal{C}_k} \|e - \mu_k\|^2
\end{equation}
where \( \mathcal{C}_k \) represents the \( k \)-th cluster, \( \mu_k \) is the centroid of \( \mathcal{C}_k \), and \( \|e - \mu_k\|^2 \) is the squared Euclidean distance between the embedding \( e \) and the cluster centroid \( \mu_k \). We then randomly select a subset of images within each cluster, to form the experience replay buffer \( \mathcal{B} \). Fine-tuning the model on this buffer \( \mathcal{B} \) ensures that the models is trained on a balanced data mix to avoid catastrophic forgetting.

\section{Experimental Results}
\label{sec:results}
In this section, we detail the experimental evaluation of \algname{}. Our goal is to demonstrate that within the proposed framework, a small, efficient object detection model for vacuum cleaner robots can effectively learn from a VLM and adapt to its operating environment over time without catastrophic forgetting. We start by creating a dataset of images representing a robotic vacuum operating in a household setting and use this data to showcase the effectiveness of language-based clustering and its advantages over vision-based clustering. We then present the results of the KD process along with our language-based continual learning approach and analyze the energy consumption over 9 virtual days of operation with one training run per day. Finally, we report the queries made to the VLM during these days, highlighting a reduction in the number of queries.

\subsection{Setup and Dataset}

We employed a TurtleBot 4 Pro robot for our experiments, a research platform derived from the Roomba vacuum cleaning robot that is equipped with an OAK-D-PRO \mbox{RGB-D} camera. The camera angle was adjusted downward to provide a closer view of the floor, thereby enhancing the visibility and clarity of both objects and surfaces. Due to the platform's lack of a dedicated GPU, we executed our fine-tuned online detector on a workstation equipped with two NVIDIA RTX 6000 GPUs and an AMD Ryzen Threadripper PRO 64-core CPU. The YOLOv8n model, serving as the backbone for our online classifier, is reputedly capable of real-time performance on edge devices such as the NVIDIA Jetson \cite{ultralyticsJetsonGuide}, making it a viable candidate for deployment on robotic vacuum systems. Processing related to the LLM, VLM, open-vocabulary object detector and the learning pipeline, was carried out on the same workstation.

We used this setup to generate a dataset of $2,500$ images capturing what a vacuum cleaner would typically observe in household environments. The dataset comprises three distinct flooring patterns and 12 different item categories. Example images of each category are presented in \autoref{fig:yolo}.

\subsection{Language-based Clustering}
To assess the effectiveness of the language-based clustering approach, we initiated our evaluation by randomly selecting a subset of data comprising 700 images. We utilized the GPT-4o VLM to obtain text descriptions and employed the \textit{text-embedding-ada-002} model for generating corresponding text embeddings. Subsequently, we applied \(k\)-means clustering to these embeddings, as detailed in \autoref{sec:approach}. For comparative purposes, we also conducted vision-based clustering using visual features extracted from ResNet50.

\begin{figure*}[htbp]
\centering
\includegraphics[clip,trim=0cm 6.6cm 0cm 0cm,width=0.85\linewidth]{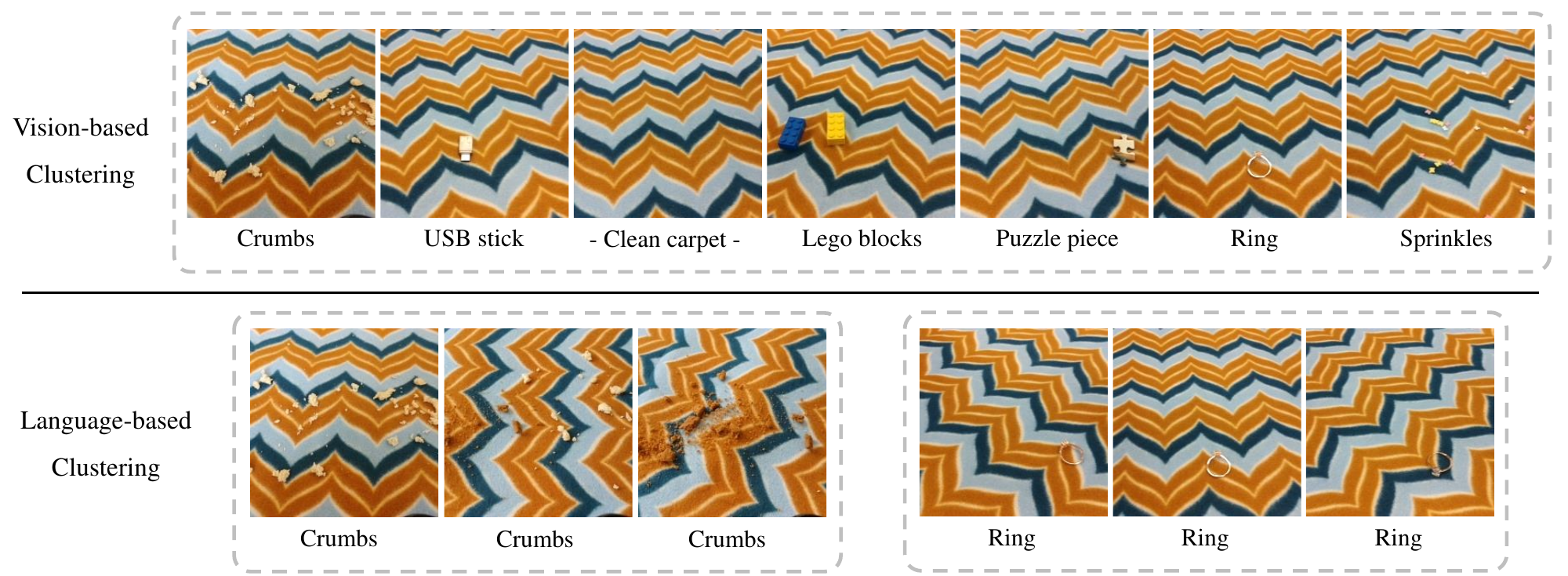}  
\caption{Sample clusters from \(k\)-means clustering. The top row displays a sample cluster from the vision-based clustering approach, while the bottom row shows two sample clusters from the language-based clustering approach.}
\label{fig:clustering}
\end{figure*}

The clustering process was executed for 20 clusters, and the mean class purity was calculated for both the vision-based and language-based clustering approaches. The mean class purity was determined using a manually defined ground truth, where images were considered to belong to the same class if both the object and the background were categorized similarly. This metric allowed us to rigorously compare the performance of the two clustering methods.
The results demonstrate a clear distinction between the two approaches: the mean class purity for language-based sampling is $93.11\%$ significantly higher than the $74.12\%$ achieved by the vision-based clustering. \autoref{fig:clustering} provides examples of clusters generated by both approaches. As illustrated, vision-based method relies heavily on background features, which leads to the incorrect grouping of small items (e.g., rings, crumbs, pet waste) into a single cluster. In contrast, language-based clustering effectively groups similar items together despite some variations in the generated labels, such as ``screws'' and ``nails'' or ``multi-color zigzag carpet'' and ``blue and orange chevron carpet''. This capability is crucial for applications like vacuuming, where the robot must differentiate between small objects (e.g., rings, crumbs) on diverse backgrounds. Based on these results, we proceed with the language-based sampling strategy for our experience replay approach.

\subsection{Knowledge Distillation and Continual Learning}
Building on these initial findings, we conducted experiments over $9$ consecutive days, with the vacuum cleaner robot visiting one of the three rooms each day. Each room featured one of the three types of flooring—gray carpet, wooden floor, and zigzag-patterned carpet—and contained $7$ different item categories. 
Common household items and dirt (e.g., crumbs) were repeated in all rooms, while less common items (e.g., screws, rings, paperclips) were present in only one or two of the rooms. We randomly selected images from the categories in the dataset and distributed them across the $9$ days. Additionally, we assumed the robot revisited each room every third day following the sequence of gray carpet, wooden floor, and zigzag carpet. This setup allowed us to investigate our algorithm's ability to adapt to new data domains while retaining previously acquired knowledge. By spacing out the visits to each room, we were able to assess how effectively the algorithm updates its model with new data without forgetting prior information—an essential factor for maintaining consistent performance in continuously changing environments.

We then conducted the experiments on the $9$ consecutive days, utilizing YOLOv8n as student model. Each day, the model classified images into actionable categories of ``to avoid'' or ``to suck''. To minimize false detections, we set a high confidence threshold of $0.9$ for accepting the model's output. Following our method described in the \autoref{sec:approach}, our framework queried the VLM for images where the model was uncertain, and the text descriptions along with bounding boxes were stored in the experience pool. We used GPT-4o as the VLM for generating image text descriptions and \mbox{OWL-ViT} for generating the bounding boxes.

To benchmark our approach, we compared it against two baseline methods: cumulative learning and naive fine-tuning. In the cumulative learning approach, the YOLOv8n model was trained from scratch each day using all the data accumulated in the experience pool up to that point. For naive fine-tuning, the YOLOv8n model was trained only on the data acquired from the previous day, but with a warm start from earlier training sessions.

To evaluate the performance of these approaches, we focused on the $F_1$ score, which offers a balanced assessment of both precision and recall. Given that our emphasis is more on accurate classification than on the precision of the bounding boxes, the $F_1$ score provides a reliable metric for evaluating the overall performance of the model. It is computed on each day on the new data of that day before training.

\begin{figure}[ht]
\centering
\includegraphics[clip,trim=0cm 0cm 0cm 0cm,width=1\linewidth]{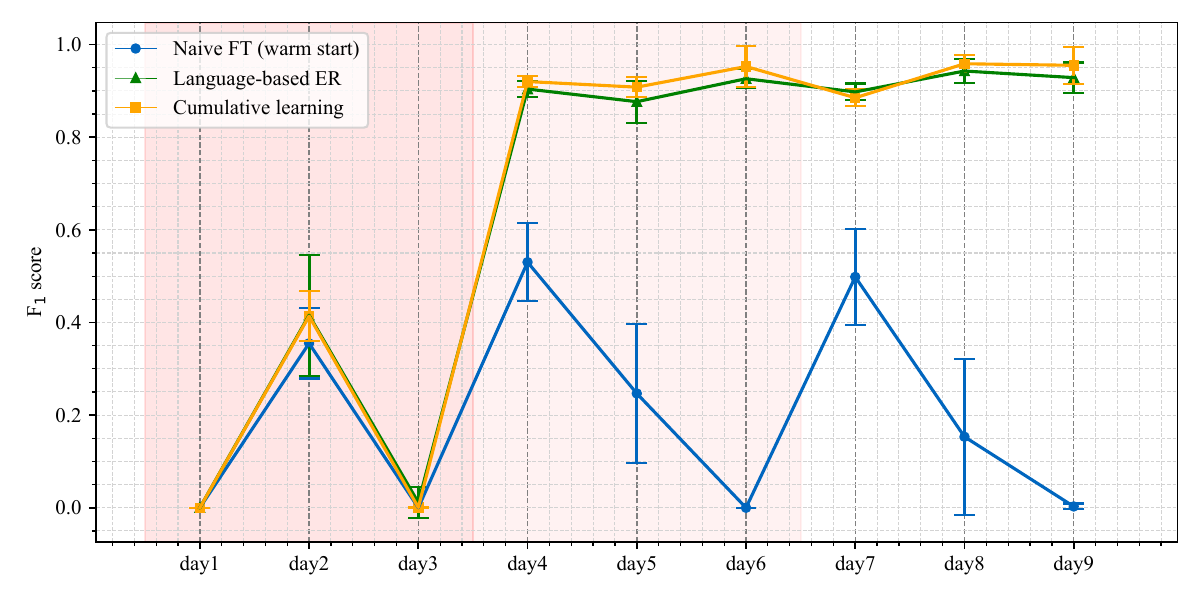}  
\caption{The $F_1$ score for naive fine-tuning, language-based experience replay and cumulative learning across $9$ consecutive days averaged over 10 runs with different random seeds. Background shadings represent 3-day intervals. The bars indicate standard deviations.}
\label{fig:F1}
\end{figure}

\autoref{fig:F1} presents the $F_1$ scores for the three methods averaged over 10 runs. We trained the model for $100$ epochs on each day. As can be seen, naive fine-tuning suffers from catastrophic forgetting, with performance periodically dropping. This behaviour is expected as the robot encounters rooms with drastically different flooring backgrounds and items over different days. However, both the language-based experience replay and cumulative learning approaches demonstrate similar performance trends, with rapid improvement after the robot has visited each room once.

\begin{figure}[ht]
\centering
\includegraphics[clip,trim=0cm 0cm 0cm 0cm,width=1\linewidth]{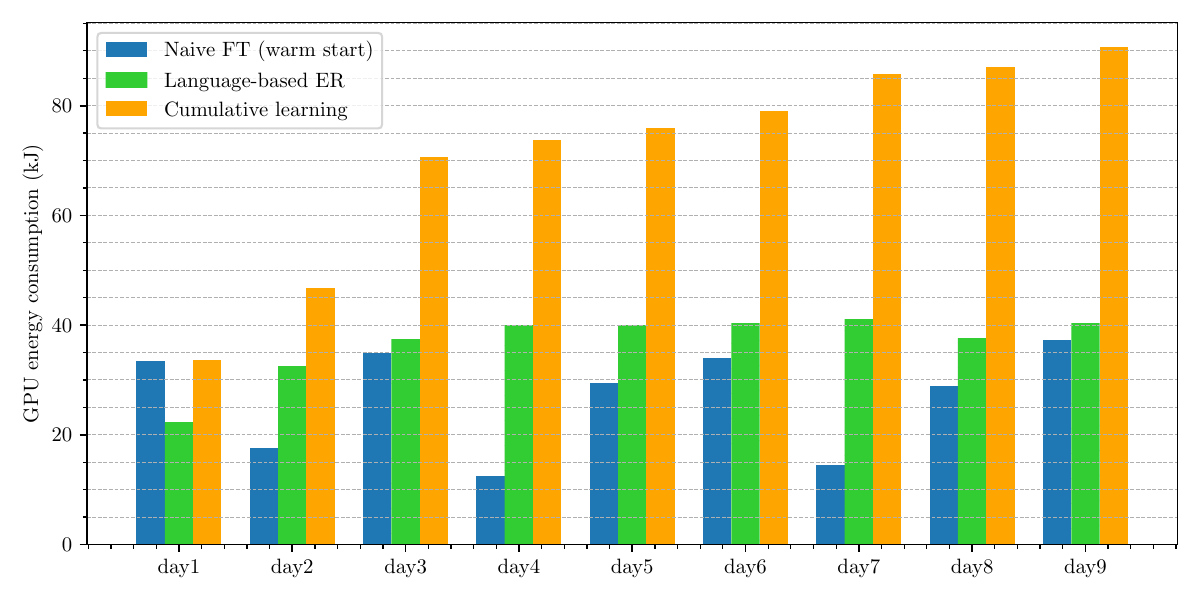}  
\caption{GPU energy consumption during training for the language-based experience replay approach, cumulative learning, and naive fine-tuning.}
\label{fig:energy}
\end{figure}

Next we analyzed the GPU energy consumption based on the power draw reported by the GPU driver over time. The results in \autoref{fig:energy} indicate energy usage for each day across the different methods. While the reported numbers are specific to our hardware setup, the relative differences in energy consumption are expected to apply also to other platforms. As illustrated in the figure, cumulative learning consumes considerably more energy due to the increasing size of the training data each day. The mean $F_1$ performance score and GPU energy consumption from day $4$ onward over the 10 runs are presented in \autoref{table:summary}. As can be seen, the language-based approach results in an $F_1$ score of $0.913$ comparable to $0.930$ achieved by cumulative learning, but reduces energy consumption by $53\%$.

\begin{figure*}[ht]
\centering
\includegraphics[clip,trim=0cm 3.5cm 0cm 0cm,width=0.9\linewidth]{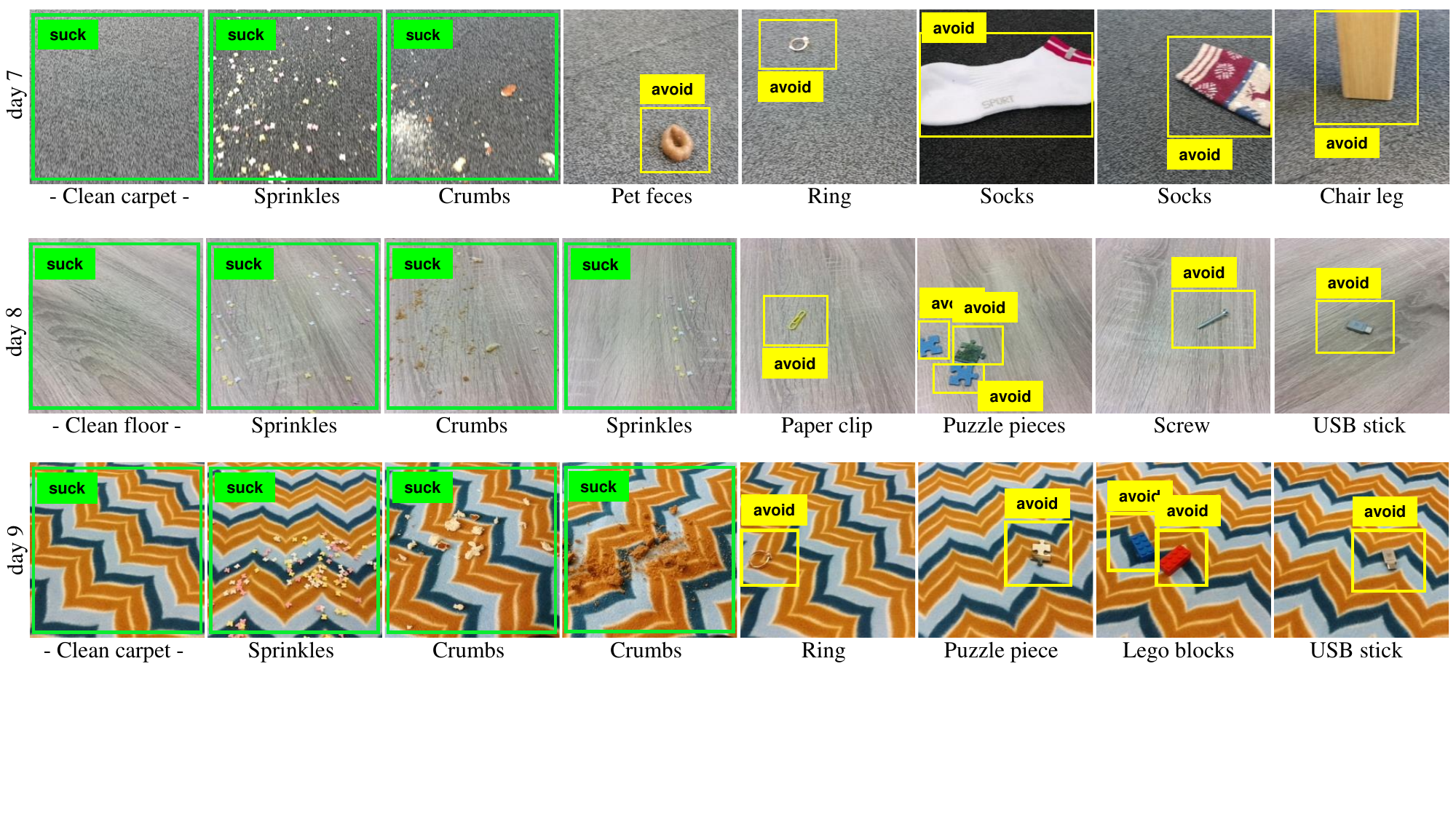}  
\caption{Results of action-based object classification with fine-tuned YOLOv8n on days $7$, $8$, and $9$. The model effectively classifies small objects even in complex backgrounds. To improve visualization, we enlarged the bounding boxes.}
\label{fig:yolo}
\end{figure*}

\begin{table}[!t]
\centering
\caption{Mean $F_1$ score and energy consumption over the last $6$ days.}
\label{table:summary}
\begin{tabular}{|>{\centering\arraybackslash}m{0.3\linewidth}||>{\centering\arraybackslash}m{0.1\linewidth}|>{\centering\arraybackslash}m{0.3\linewidth}|}
\hline
\renewcommand{\arraystretch}{1.5}
 & Mean $F_1$ & Mean GPU energy consumption (kJ) \\
\hline
\hline
Naive fine-tuning (warm start) & 0.239 & \textbf{26.1} \\
\hline
Language-based experience replay & \textbf{0.913} & \textbf{39.3} \\
\hline
Cumulative learning & \textbf{0.930} & 83.6 \\
\hline
\end{tabular}
\end{table}

\begin{figure}[h]
\centering
\includegraphics[clip,trim=0cm 0cm 0cm 0cm,width=1\linewidth]{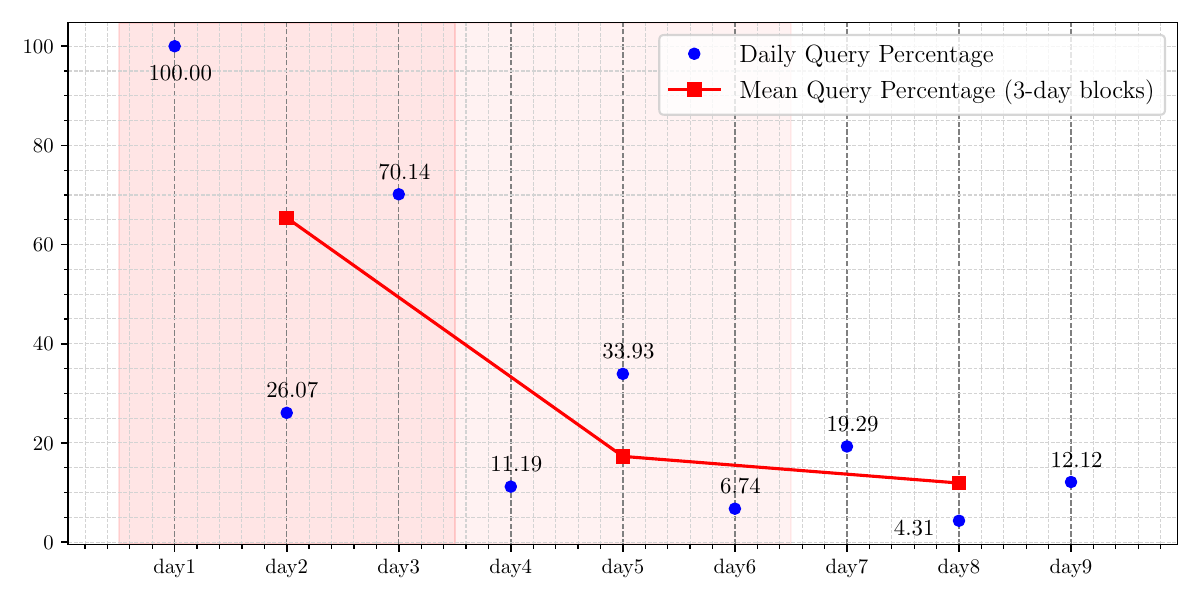}  
\caption{Percentage of images queried to the VLM over nine consecutive days, with the mean query percentage indicated for each three-day interval.}
\label{fig:queries}
\end{figure}

Finally, the object classification results for days $7$, $8$ and $9$, using the language-based experience replay approach are depicted in \autoref{fig:yolo}. The bounding boxes show the class categories that the fine-tuned YOLO model detected for each item. As shown in the figure, our method effectively performs object detection, even for small items and on complex flooring patterns. Additionally, \autoref{fig:queries} illustrates the percentage of images queried to the VLM each day using the language-based experience replay method. The red line represents the mean query percentage over three-day blocks. The decreasing trend in the mean query percentage supports our claim that knowledge from the VLM is gradually being distilled into the smaller model, YOLOv8n.

\section{Conclusion and Future Work}
In this paper, we proposed \algname{}, a novel framework for enhancing the capabilities of smart vacuums in dynamic real-world environments. By leveraging knowledge distillation, we transfer the knowledge from a VLM to a smaller, more efficient model increasing the system's efficiency. We also introduce a novel language-based continual learning approach to mitigate catastrophic forgetting.

One interesting direction for future work is to investigate the long-term performance of our approach. According to recent studies \cite{plasticity}, the learning capacity of deep-learning models declines when trained on new data over extended periods. Wang~\etal~\cite{plasticity}, proposed to tackle this phenomenon by continuously reinitializing a fraction of the less-active neurons. They report that this approach helps deep-learning models to indefinitely maintain their plasticity and keep learning from new streams of data. This is particularly relevant to our approach, as a robotic vacuum cleaner needs to operate in dynamic environments and continuously learn from new data. 

Another potential approach for future work is to deploy an out-of-distribution (OOD) detection algorithm in our framework. This addition could prevent the vacuum cleaner from making confident wrong guesses when faced with unfamiliar objects or backgrounds that deviate from its training data. Finally, it would be interesting to include a broader range of objects and backgrounds in the experiments, as well as to conduct them over a longer period of time. This expansion could help provide more insights about the performance of our algorithm when faced with diverse real-world scenarios.

\label{sec:conclusion}



\bibliographystyle{IEEEtran}
\bibliography{sources.bib}

\end{document}